\documentclass[11pt,a4paper]{article}

\usepackage{polyu-vclab}

\usepackage{lipsum}                
\usepackage{amsmath}
\usepackage{booktabs}
\usepackage{multirow}
\usepackage[numbers,sort&compress]{natbib}
\usepackage{subcaption}
\usepackage{wrapfig}
\usepackage{algorithm, algorithmic}
\usepackage{bm}

\newcommand{\figdashsep}{%
  \raisebox{-14mm}{%
    \makebox[0pt][c]{%
      \vbox to 33mm{%
        \offinterlineskip
        \leaders\vbox{\hbox{\textcolor{gray}{\rule{0.6pt}{2.2mm}}}\vskip1.8mm}\vfill
      }%
    }%
  }%
}

\setEyebrow{PolyU VCLab\,\textbullet\,Preprint 2026}

\setReportTag{Visual Computing Lab\,\textperiodcentered\,The Hong Kong Polytechnic University}

\papertitle{Weighted Reverse Convolution for Feature Upsampling}

\paperauthors{%
  Wentong Li\equalmark\affilmark{1}\quad
  Zhiyuan Qi\equalmark\affilmark{1,2}\quad
  Zichen Zhao\affilmark{1}\quad
  Kai Zhang\affilmark{3}\quad
  Lei Zhang\correspondmark\affilmark{2}
}

\paperaffil{%
  \affilmark{1}\,Nanjing University of Aeronautics and Astronautics\quad
  \affilmark{2}\,The Hong Kong Polytechnic University\quad
  \affilmark{3}\,Nanjing University
}

\papernotes{%
  \equalmark\,Equal contribution (\href{mailto:liwentong95@gmail.com,zhi-yuan.qi@polyu.edu.hk}{liwentong95@gmail.com,zhi-yuan.qi@polyu.edu.hk}). \quad
  \correspondmark\,Corresponding author (\href{mailto:cslzhang@comp.polyu.edu.hk}{cslzhang@comp.polyu.edu.hk}).%
}

\paperbadges{%
  \centering{\vclabbadgesolid{arXiv:2026.17472}{https://arxiv.org/abs/2605.17472}\;%
  \vclabbadgesolid{Code}{https://github.com/PolyU-VCLab/WRC}\;}
}

\begin{document}

\maketitleVCLab

\begin{vclabAbstract}
\noindent\textbf{Abstract.}\;
Pre-trained vision foundation models (VFMs) provide strong semantic representations, yet their patch-level features are inherently coarse, limiting their effectiveness on tasks requiring  fine-grained localization, dense prediction, and point-wise correspondence.
In this work, we revisit feature upsampling for VFMs from the perspective of \textbf{\textit{inverse problem}} and propose Weighted Reverse Convolution (WRC), a spatially adaptive inverse operator for densifying high-level visual descriptors.
Specifically, we formulate feature upsampling as a weighted Tikhonov-regularized least-squares problem, where spatially varying weights modulate both data fidelity and prior strength at each spatial location.
This allows WRC to adapt the reconstruction to spatially varying feature characteristics, thereby preserving critical structures while mitigating over-smoothing.
Moreover, WRC retains an efficient, fully differentiable closed-form FFT solution, making it a practical drop-in upsampling operator.
Integrated into a lightweight self-supervised densification framework, WRC consistently improves dense feature quality across various downstream benchmarks, including segmentation, depth estimation, video object segmentation, object discovery, and keypoint correspondence,
while maintaining high computational efficiency.
\end{vclabAbstract}

\section{Introduction}
Pre-trained vision foundation models (VFMs) have emerged as the primary source of transferable visual representations in computer vision~\cite{liu2023visual,li2025mini, el2024scalable,yuan2024osprey,bolya2026perception}. Models such as DINO~\cite{caron2021emerging,oquab2023dinov2,simeoni2025dinov3} and CLIP~\cite{clip,tschannen2025siglip}, trained on a large scale, provide rich semantic or language-aligned features generalizable across various downstream tasks.
However, most VFMs tokenize an image into a coarse patch grid (\textit{e.g.}, with stride 14/16), resulting in spatially low-resolution feature maps.
This coarseness becomes a practical bottleneck for tasks that require precise localization, fine spatial delineation, and reliable point-wise correspondence.

A straightforward solution is to increase the token density, either by feeding higher-resolution inputs or by reducing the patch size. However, both options are expensive: self-attention scales quadratically with token count, and architectural changes often require large-scale re-training or substantial fine-tuning. Moreover, such approaches may introduce artifacts into feature maps, ultimately degrading performance~\cite{yang2024denoising,fan2024vitar}.
These limitations motivate a promising alternative solution: feature upsampling for pre-trained VFMs, where the backbone is frozen and a lightweight module is employed to reconstruct higher-resolution features that are better suited to downstream tasks.
An effective upsampler should satisfy several key requirements: (\textit{i}) preserve geometric fidelity, so that boundaries and point-wise correspondences can be accurately represented;  (\textit{ii}) maintain semantic-space fidelity, so that downstream heads can be transferred without substantial re-adaptation; (\textit{iii}) remain efficient and plug-and-play, so that it can be seamless integrated with different VFMs. Recent approaches have explored different routes, including image-guided decoders \cite{featup,loftup}, implicit coordinate-based prediction \cite{featup}, and cross-attention interpolation \cite{anyup,jafar}. Although these methods achieve strong performance, they typically face a trade-off 
between fine spatial detail recovery 
and inference efficiency.

\begin{figure*}[t]
  \vskip 0.15in
  \begin{center}    
  \hfill 
    \begin{subfigure}[c]{0.74 \linewidth}
        \centering
        \includegraphics[width=1\linewidth]{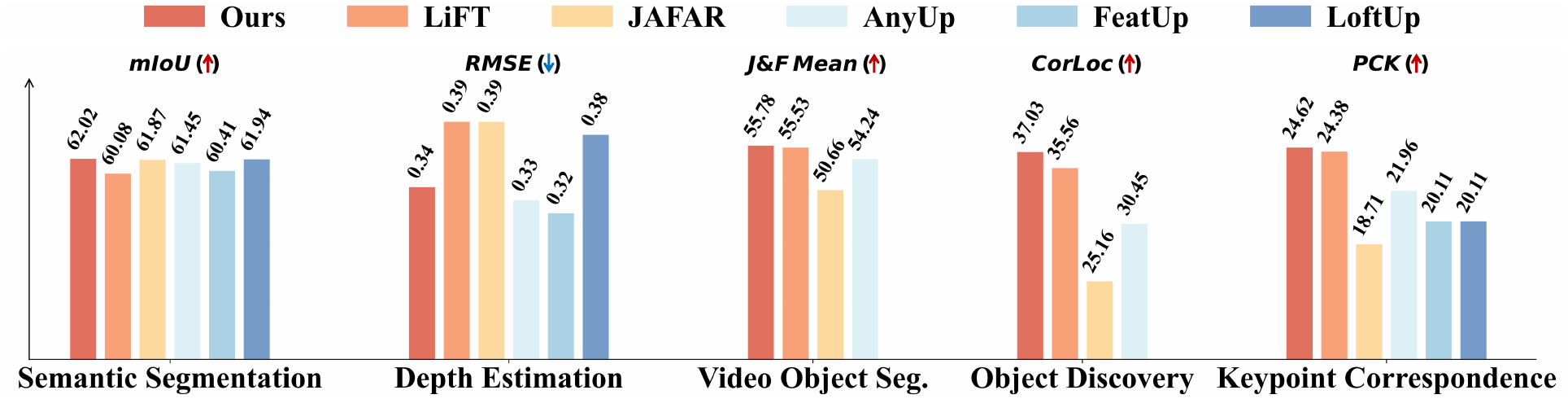}
        \caption{}
        \label{fig: intro_a}
    \end{subfigure}
    \hfill
    \hspace{-1mm}\figdashsep\hspace{0.2mm}
    \hfill
    \begin{subfigure}[c]{0.25 \linewidth}
        \centering
        \includegraphics[width=1\linewidth]{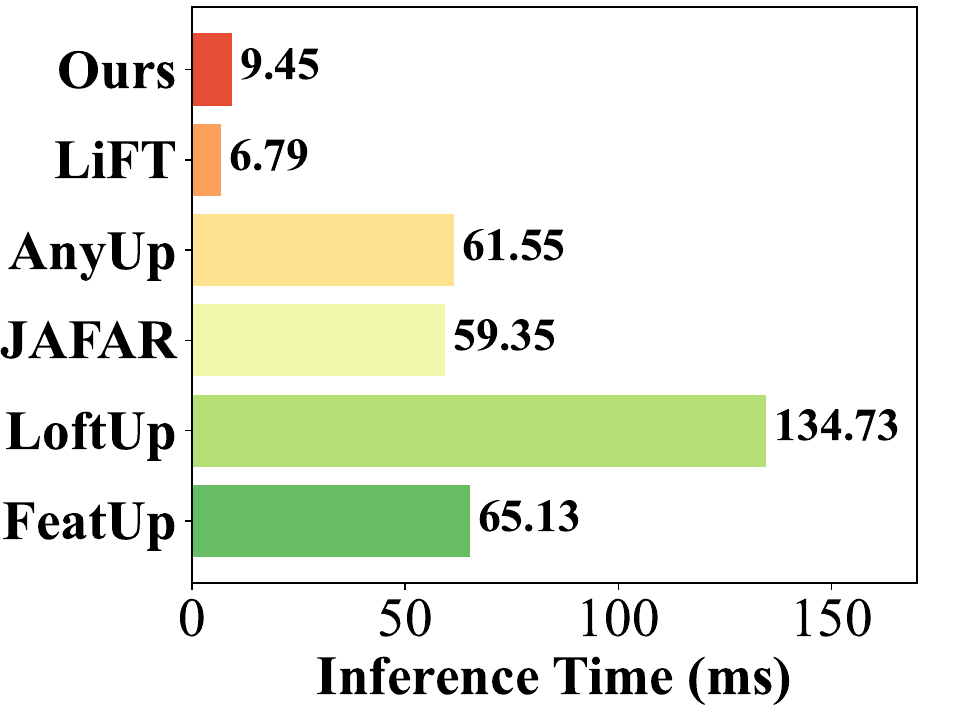}
        \caption{}
        \label{fig: intro_b}
    \end{subfigure}
    \caption{Quantitative Evaluation and Efficiency Comparison. (a) Average performance on linear probing semantic segmentation
    (mIoU), 
    linear probing depth estimation (RMSE), video object segmentation ($\mathcal{J}\&\mathcal{F}$ mean), object discovery (CorLoc), and keypoint correspondence (PCK). We compare our method with recent feature upsampling approaches, including AnyUp~\cite{anyup}, JAFAR~\cite{jafar},  LoftUp~\cite{loftup}, FeatUp~\cite{featup}, and LiFT~\cite{lift}. (b) Wall-clock inference time (ms) measured under the same hardware and input setting.}
    \vspace{-1mm}
    \label{fig:intro}
  \end{center}
\end{figure*}

To address this challenge, we explore a novel perspective by treating feature upsampling as an \textbf{\textit{inverse problem}}.  Instead of learning an unconstrained decoder, we seek to restore high-resolution feature maps from their  observed low-resolution counterparts.
Reverse convolution~\cite{conversenet} provides a principled foundation for this formulation by casting low-level image recovery as a regularized inverse problem with an efficient closed-form solution computable via FF. 
This perspective is also well aligned with the structure of ViT-based VFMs, whose self-attention maps often exhibit approximately block-circulant patterns, suggesting a convolution-like and Fourier-friendly structure~\cite{han2026vision}.
However, directly applying such image restoration-oriented inverses to VFM features is suboptimal. Dense transformer features in VFMs are different from the restoration features in natural images:
they encode high-level semantics structures across space and channels, and 
uniform least-squares fidelity with a global regularizer tends to over-smooth discriminative cues that downstream tasks rely on.

In this work, we propose Weighted Reverse Convolution (WRC), a spatially adaptive inverse operator tailored for feature upsampling in pre-trained VFMs. In contrast to standard reverse convolution, WRC formulates feature upsampling as a spatially weighted, Tikhonov-regularized inverse problem, where data fidelity and prior regularization are adaptively modulated at each location.
Specifically, we introduce positive, input-adaptive per-location weights into both the reconstruction and regularization terms, allowing the operator to adjust the strength of data fitting and prior enforcement.
This design helps preserve correspondence-critical structures and mitigate over-smoothing in ambiguous regions. Moreover, we drive an efficient closed-form FFT solution, making WRC practical as a drop-in upsampling operator.

We integrate WRC into a lightweight  densification framework for frozen VFM features. A lightweight CNN enriches low-resolution descriptors with image details, and WRC reconstructs denser features with improved spatial precision and semantic stability, guided by global reconstruction, local crop reconstruction, and self-consistency losses.
As shown in Fig.~\ref{fig:intro},
extensive experiments across diverse downstream benchmarks, including linear probing segmentation and depth estimation, video object segmentation, unsupervised object discovery, and keypoint correspondence, demonstrate the consistent improvement of WRC over recent feature upsampling approaches while maintaining high computational efficiency.

Our contributions are summarized as follows:

\begin{itemize}
  \item We revisit feature upsampling for VFMs from an inverse-problem perspective, providing a principled formulation to recover dense high-level representations from coarse features.
  \item  We propose Weighted Reverse Convolution (WRC), a spatially adaptive inverse upsampler that modulates both data fidelity and Tikhonov regularization with input-adaptive weights, enabling finer and more stable descriptors reconstruction.
  \item We derive an efficient closed-form FFT solution for WRC, making it practical as a fully differentiable drop-in upsampling operator within lightweight densification frameworks. 
  \item We demonstrate that WRC consistently improves dense feature quality across diverse downstream tasks while maintaining high computational efficiency.
\end{itemize}

\section{Related Work}

\subsection{Feature Upsampling}
Feature upsampling for VFMs aims to increase the spatial resolution of intermediate representations to better support dense prediction and correspondence, without incurring the quadratic cost by increasing the token count of pre-trained ViT models~\cite{li2025tokenpacker,han2026vision}. An effective upsampler should not only recover fine spatial detail, but also preserve the semantics and distribution of the original feature space. General feature upsampling methods~\cite{Wang2019CARAFECR, Lu2022SAPASP, Liu2023LearningTU} directly operate on the low-resolution feature using interpolation~\cite{keys2003cubic} or learnable upsampling blocks such as transposed\&resized  convolution~\cite{noh2015learning}. These approaches are efficient and modular, but they often struggle to recover sharp boundaries and correspondence-critical details.
Recent work increasingly conditions upsampling on the high-resolution input image to inject fine-grained structure. LiFT~\cite{lift} employs a transposed decoder to densify frozen ViT features. FeatUp~\cite{kopf2007joint} combines an image-guided operator based on Joint Bilateral Upsampling (JBU), and employs an implicit formulation to parameterize high-resolution features.
LoftUp~\cite{loftup} and JAFAR~\cite{jafar} instead pursue coordinate-/attention-based reconstruction to mitigate the cumulative artifacts from multi-stage upsampling. 
AnyUp~\cite{anyup} argues that upsampled features should remain in the same space as the original features, and formalizes this route as feature space preservation.

\subsection{Deconvolution}
In deep networks, \textit{deconvolution} often refers to transposed convolution~\cite{noh2015learning}, which is not the true inverse of convolution; it can be interpreted as inserting zeros between feature-map elements, followed by a standard convolution. In classical signal and image processing, deconvolution refers to inverting a known convolutional operation to recover a sharp signal. 
Wiener deconvolution~\cite{wiener1949extrapolation} provides a canonical closed-form approach based on inverse filtering with regularization when the blur kernel is known, while multi-step solutions such as deep unfolding~\cite{zhang2017beyond,mou2022deep}, unroll iterative optimization into trainable networks. 
However, these methods are typically designed for natural images and cannot be directly translated to the high-dimensional semantic feature maps produced by vision foundation models.
Recently, 
reverse convolution ~\cite{conversenet} bridges these perspectives by formulating inversion as a regularized least-squares problem, enabling a closed-form operator that enforces data consistency. Its FFT-based solution for inverting depthwise convolution provides an efficient alternative to heuristic upsampling blocks and provides a natural foundation for our weighted reverse convolution design.

\section{Preliminary}
\subsection{Reverse Convolution}
\textbf{Transposed Convolution} is a standard upsampling operator in encoder-decoder architectures~\cite{noh2015learning}.
Consider a single-channel feature map $\mathbf{X} \in \mathbb{R}^{H\times W} $ and a convolution kernel $\mathbf{K} \in \mathbb{R}^{k_H\times k_W} $, 
where $H$ and $W$ denote the spatial height and width of the feature map, and $k_H$ and $k_W$ denote the kernel size.
A corresponding strided convolution can be formulated as:
\begin{equation}
    \mathbf{Y} = \left( \mathbf{X} \otimes \mathbf{K} \right)\downarrow_s,
    \label{eq: convolution}
\end{equation}
where  $\otimes$ denotes the convolution operator, $\downarrow_s$ denotes downsampling  with stride $s$. The output feature map is $\mathbf{Y} \in \mathbb{R}^{H'\times W'}$, with  $H'$ and $W'$ determined by the kernel size, stride and padding.
Transposed convolution aims to recover an upsampled feature  $\widehat{\mathbf{X}}$ from  low-resolution $\mathbf{Y}$.
This is an ill-posed problem that serves as a learnable heuristic upsampling rather than a mathematically exact inverse of Eq.~\eqref{eq: convolution}.

\begin{figure*}[t]
  \vspace{-2mm}
  \begin{center}
    \centerline{\includegraphics[width=1\linewidth]{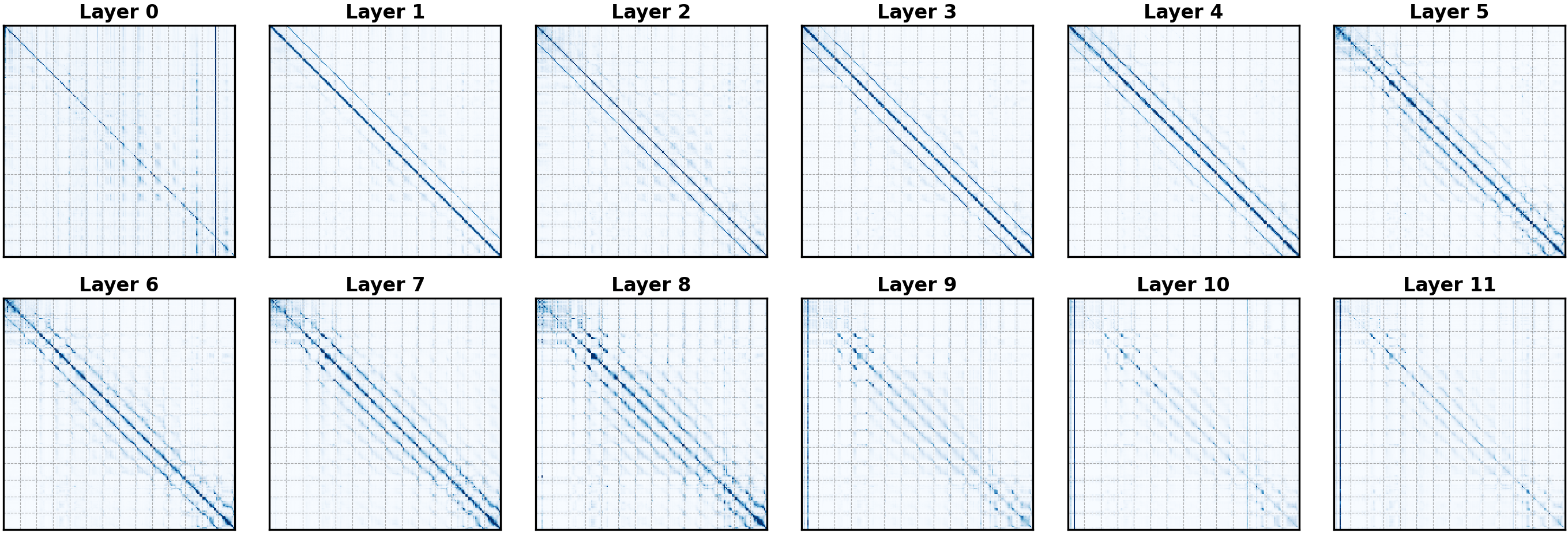}}
    \caption{Per-layer attention maps averaged over all heads in DINOv2~\cite{dinov2}. 
The self-attention modules in Vision Transformer naturally exhibit near BCCB patterns. Additional visualizations for DINOv3~\cite{simeoni2025dinov3} are provided in \textbf{Appendix~\ref{sec:bccb}}, where similar patterns can be observed.}
    \label{fig:attention}
    \vspace{-6.8mm}
  \end{center}
\end{figure*}

\noindent\textbf{Reverse Convolution}~\cite{conversenet} explicitly models feature upsampling as a regularized inverse problem.
It explicitly recovers a high-resolution feature map $\mathbf{X}$ by minimizing the discrepancy between the observed output $\mathbf{Y}$ and the forward model $\left( \mathbf{X} \otimes \mathbf{K} \right)\downarrow_s$, defining the optimization problem as:
\begin{equation}
    \mathbf{X}^* = \underset{\mathbf{X}}{\arg\min}
    \left\| \mathbf{Y} - \left( \mathbf{X} \otimes \mathbf{K} \right) \downarrow_s \right\|_F^2
    + \lambda \left\| \mathbf{X} - \mathbf{X}_0 \right\|_F^2,
    \label{eq: pre converse2d}
\end{equation}
where $\lambda > 0$ controls the regularization strength. $\mathbf{X}_{0}$ is an initial estimate of $\mathbf{X}$ such as a zero map or an interpolation-based upsampling of $\mathbf{Y}$. $\left\| \cdot \right\|_F$ denotes the Frobenius norm.

Under the assumption of circular boundary conditions, reverse convolution drives a closed-form solution to Eq.~\eqref{eq: pre converse2d}:
\begin{equation}
    \mathbf{X}^* = \mathcal{F}^{-1} \left( \frac{1}{\lambda} {\left(\mathbf{L} - \overline{\mathcal{F}_\mathbf{K}} \odot_s \frac{( \mathcal{F}_\mathbf{K} \odot \mathbf{L}) \Downarrow_s }{|\mathcal{F}_\mathbf{K} |^2\Downarrow_s + \lambda } \right)} \right),
    \label{eq: converse2d}
\end{equation}
where $\mathbf{L} = \overline{\mathcal{F}_\mathbf{K}} \mathcal{F}_{ \mathbf{Y} \uparrow_s} + \lambda \mathcal{F}_{\mathbf{X}_0}$. Here,   
$\mathcal{F}(\cdot)$ denotes the fast Fourier transform (FFT) and $\mathcal{F}^{-1}(\cdot)$ denotes its inverse;
$\overline{\mathcal{F}_{\mathbf{K}}}$ is the complex conjugate of $\mathcal{F}_\mathbf{K}$; $\odot_s$ is element-wise multiplication applied to $s\times s$ distinct blocks; $\Downarrow_s$ is the distinct block downsampling operator that averages each $s\times s$ block; $\uparrow_s$ is the standard $s$-fold upsampling operator that inserts zeros.
This operator can serve as a modular block within deep
neural networks, 
enhancing their capacity to reconstruct complex spatial structures, which has achieved strong performance on image restoration tasks \cite{conversenet}.

\subsection{BCCB Patterns in Attention}

Recent evidence~\cite{han2026vision} shows that self-attention in Vision Transformers often learns near Block Circulant with Circulant Blocks (BCCB) patterns without explicitly imposing such constraints. Intuitively, this means that the attention distribution of one query patch is approximately a shifted version of the distributions of nearby query patches, resembling the translation-sharing behavior of convolution.

We observe a similar pattern in DINOv2~\cite{dinov2}, as shown in Fig.~\ref{fig:attention}. By visualizing its patch-to-patch attention maps, we find that many heads exhibit clear block-wise circular structures and can be approximated by their nearest BCCB matrices. This provides structural motivation for applying reverse convolution to VFM (\textit{e.g.}, DINOv2~\cite{dinov2}, DINOv3~\cite{simeoni2025dinov3}) features: the backbone attention itself behaves in a convolution-like, Fourier-friendly manner. At the same time, the residual between the raw attention and its BCCB approximation suggests that this structure is spatially varying, which further motivates the weighted formulation in our WRC.

\section{Method}
We present a feature upsampling method tailored to ViT-based VMFs, aiming to recover high-resolution, dense ViT descriptors that are spatially precise for  downstream
tasks that require fine-grained spatial understanding,  while keeping the pre-trained backbone frozen. We build upon a lightweight densification framework, and 
introduce a novel feature upsampler, \textbf{Weighted Reverse Convolution (WRC)}. In the following section, we first (\textit{i}) introduce the overall framework, then (\textit{ii}) formulate WRC as a weighted, Tikhonov-regularized inverse problem and provide an efficient closed-form solution, and finally (\textit{iii}) describe the training objectives that combine global and local self-supervision.

\begin{figure*}[t]
  \begin{center}
  \centerline{\includegraphics[width=1\linewidth]{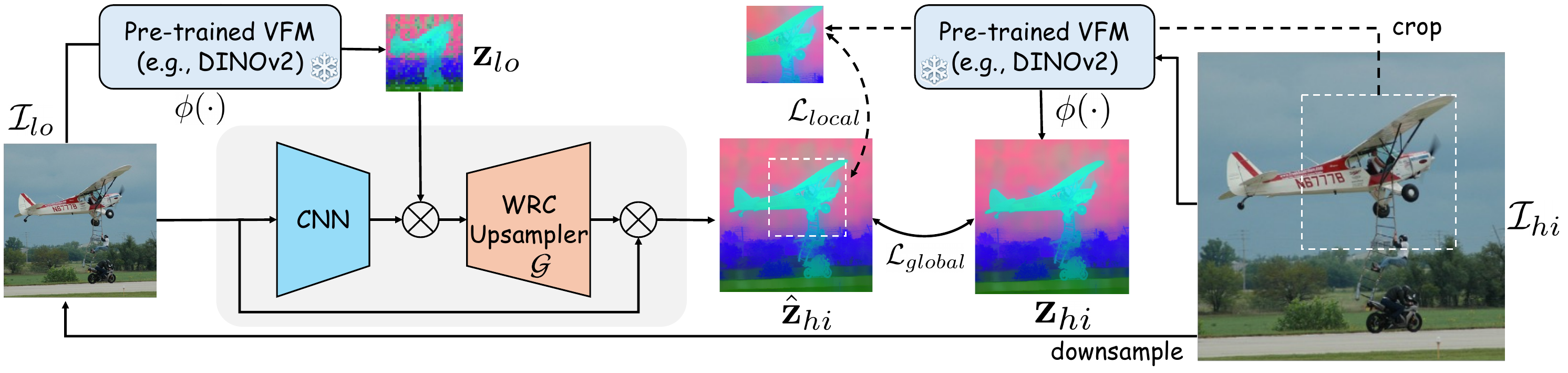}}
    \caption{\textbf{Overview of Architecture.} Given a high-resolution image, a frozen VFM extracts supervision features, while a lightweight CNN encodes image details to fuse with the low-resolution VFM features. The fused representation is then upsampled by our WRC upsampler to produce dense descriptors.
    Training is self-supervised via global reconstruction  $\mathcal{L}_{global}$ and local crop reconstruction $\mathcal{L}_{local}$.
    }
    \label{fig:method}
  \end{center}
  \vspace{-6.5mm}
\end{figure*}

\subsection{Framework Overview}
As illustrated in Fig.~\ref{fig:method}, given a high-resolution RGB image \(\mathbf{\mathcal{I}}_{hi} \in \mathbb{R}^{M \times N \times 3}\), a pre-trained VFM encoder \(\phi(\cdot)\) is used to produce a feature tensor at a lower spatial resolution. A feature upsampling operator \(\mathcal{G}\) lifts these low-resolution features to a higher-resolution feature map to better support dense, pixel-wise or point-wise reasoning tasks.  Let $\mathbf{\mathcal{I}}_{lo}$ denote a 2$\times$ downsampled version of $\mathbf{\mathcal{I}}_{hi}$. 
Specifically, a frozen pre-trained backbone \(\phi(\cdot)\) (\textit{e.g.}, DINOv2) extracts features from both $\mathbf{\mathcal{I}}_{hi}$ and $\mathbf{\mathcal{I}}_{lo}$. The framework encodes image details from $\mathbf{\mathcal{I}}_{lo}$ using  downsampling CNN and fuses them with the backbone feature $\mathbf{z}_{lo}$ to form a joint representation. 
This fused feature is subsequently passed to the upsampling operation. 
Our WRC upsampler is designed to better preserve the discriminative structure of high-level descriptors while recovering spatial details. The entire model is trained in a self-supervised manner by matching the predicted feature $\hat{\mathbf{z}}_{hi}$ to the teacher feature $\mathbf{z}_{hi}$ extracted from $\mathbf{\mathcal{I}}_{hi}$ using global and local reconstruction losses.

\subsection{Weighted Reverse Convolution as Upsampler}

\subsubsection{Problem Formulation}
A key limitation of directly applying restoration-style inverse objectives, \textit{e.g.}, uniform MSE to high-level descriptors, is that they implicitly treat all spatial locations and channels as equally important.
In contrast, dense ViT descriptors are spatial heterogeneous: boundary regions, thin structures, and ambiguous textures contribute disproportionately to downstream dense prediction and correspondence performance. 
Enforcing uniform fidelity across all positions can bias the solution toward over-smoothed features, diminishing the discriminative cues that high-level downstream tasks rely on.

To address this, we introduce a weighted data-fidelity term and formulate feature restoration as a weighted least-square optimization problem to adaptively preserve the discriminative structure of the feature map:
\begin{equation}
    \mathbf{X}^* = \underset{\mathbf{X}}{\arg\min} \left\| \mathbf{W} \left( \mathbf{Y} - \left( \mathbf{X} \otimes \mathbf{K} \right) \downarrow_s \right) \right\|_F^2,
    \label{eq: weighted least-squares}
\end{equation}
where $\mathbf{W} \in \mathbb{R}^{\frac{M}{2}\times \frac{N}{2}}$ contains positive weights. 
$\mathbf{W}$ modulates the contribution of each spatial location to the inversion, allowing the model to down-weight uncertain regions while emphasizing correspondence-critical structures.

\subsubsection{Semantic Prior via Tikhonov Regularization}
Converse2D~\cite{conversenet} stabilizes inversion with an isotropic $\ell_2$ penalty, which can be overly restrictive for high-level semantic descriptors: it penalizes all feature magnitudes uniformly and could oversmooth spatial details that are important for semantic delineation. We generalize the regularization to a Tikhonov form~\cite{tikhonov1995numerical}:
\begin{equation}
    \left\| \bm{\Gamma} (\mathbf{X} - \mathbf{X}_0) \right\|_F^2 ,
\end{equation}
where $\bm{\Gamma}$ is a data-driven adaptive matrix for linear operator and $\mathbf{X}_0$ is a prior estimate of  $\mathbf{X}$. When $\bm{\Gamma} = \sqrt{\lambda}\mathbf{I}$,  this reduces to the standard $\ell_2$ regularization. 
This generalization allows the inversion to dynamically reconstruct the spatial semantics with reference to the prior $\mathbf{X}_0$.

By jointly introducing a weighted data-fidelity term and a weighted Tikhonov prior, we arrive at the WRC objective:
\begin{equation}
    \mathbf{X}^* = \underset{\mathbf{X}}{\arg\min} \left\| \mathbf{W} \left( \mathbf{Y} - \left( \mathbf{X} \otimes \mathbf{K} \right) \downarrow_s \right) \right\|_F^2 
    + \left\| \mathbf{W}_\lambda \left( \mathbf{X} - \mathbf{X}_0 \right) \right\|_F^2,
    \label{eq: weighted reverse convolution}
\end{equation}
where $\mathbf{W}_\lambda \in \mathbb{R}^{M\times N}$ also contains positive weights. 

\subsubsection{Closed-form Solution}
Following~\cite{zhao2016fast}, we vectorize $\mathbf{X}$ and $\mathbf{Y}$ under the assumption of circular boundary conditions to simplify the problem solving process. By solving the vectorized optimization problem and reverting the solution to its matrix representation, we derive the closed-form solution for $\mathbf{X}$:
\begin{small}
\begin{equation}
\label{eq: wrc closed form1}
\mathbf{X}^* = \mathcal{F}^{-1}\Biggl[
\frac{1}{\lvert \mathbf{W}_{\lambda} \rvert^2}
\Bigl(
\mathbf{L}'-\overline{\mathcal{F}_{\mathbf{K}}}\odot_s\,\mathbf{Q}
\Bigr)
\Biggr], 
\mathbf{Q}=
\frac{
\lvert \mathbf{W} \rvert^2 \odot
\Bigl( \bigl( \mathcal{F}_{\mathbf{K}} \odot \mathbf{L'} \bigr)\Downarrow_s \Bigr)
}{
\lvert \mathbf{W} \rvert^2 \odot
\Bigl(\lvert \mathcal{F}_{\mathbf{K}} \rvert^2 \Downarrow_s \Bigr)
+\lvert \mathbf{W}_{\lambda} \rvert^2 \Downarrow_s
},
\end{equation}
\end{small}where $\mathbf{L'} = \overline{\mathcal{F}_{\mathbf{K}}} \mathcal{F}_{(\lvert\mathbf{W}\rvert^{2} \odot \mathbf{Y})\uparrow_s} + \lvert \mathbf{W}_{\lambda}\rvert^{2} \odot \mathcal{F}_{\mathbf{X}_0}$. 
This formulation is a strict generalization of  Converse2D. In particular, when $\mathbf{W} = \mathbf{I}$ and $\mathbf{W}_\lambda = \sqrt{\lambda} \mathbf{I}$, Eq.~\eqref{eq: wrc closed form1} reduces to the Converse2D solution (Eq.~\eqref{eq: converse2d}). When $s=1$, it further simplifies to the standard inverse filtering form:
\begin{equation}
    \mathbf{X}^* = \mathcal{F}^{-1}\left( \frac{ \overline{\mathcal{F}_\mathbf{K}} \odot \mathcal{F}_\mathbf{Y} + \lambda \mathcal{F}_{\mathbf{X}_0} }{ \lvert \mathcal{F}_\mathbf{K} \rvert^{2} + \lambda } \right).
\end{equation}
Overall, our WRC provides a more general and flexible inverse formulation by allowing spatially varying fidelity and regularization weights, while retaining an efficient FFT-based closed-form solution. \textit{\textbf{The detailed derivation process can be found in Appendix~\ref{sec:proof}}}.

\subsubsection{Implementation of WRC} 
\label{sec: implementation of WRC}

\noindent {\bf Weights Initialization and Parameterization.} 
WRC introduces two weights $\mathbf{W}$ and $\mathbf{W}_\lambda$, which govern the data-fidelity and regularization terms, respectively. Inspired by dynamic parameterization~\cite{jia2016dynamic}, we generate both weights dynamically using lightweight convolutional predictors conditioned on the input features, rather than learning a single global weight matrix. This design makes the weights input-adaptive and improves robustness across diverse inputs. To ensure positivity and numerical stability, we apply monotone re-parameterizations. Empirically, $log1p$ provides strong numerical stability and performance in our setting.
A PyTorch-like implementation is provided in Algorithm~\ref{alg: wrc} of \textbf{Appendix~\ref{sec:algo}}.

\noindent {\bf CUDA Implementation.} 
To make WRC more practical as a drop-in upsampling operator, we implement its closed-form solver efficiently on GPU. The computation is dominated by batched FFTs and element-wise complex arithmetic, which map well to CUDA primitives. The entire operator is differentiable and supports end-to-end training via autograd, while maintaining low overhead compared to previous upsampling blocks.

\subsection{Training Objective}
We train the upsampler by integrating global self-supervision and local self-supervision, which together provide coarse-to-fine constraints.

\noindent {\bf Global Self-supervision.}
We use the pre-trained VFM encoder $\phi$ to extract $\mathbf{z}_{lo}=\phi(\mathbf{\mathcal{I}}_{lo})$ from the $2\times$ downsampled image $\mathbf{\mathcal{I}}_{lo}$ and predict $\hat{\mathbf{z}}_{hi}=\mathcal{G}(\mathbf{z}_{lo},\mathbf{\mathcal{I}}_{lo})$ by upsampler $\mathcal{G}$. We then align $\hat{\mathbf{z}}_{hi}$ with the teacher feature $\mathbf{z}_{hi}=\phi(\mathbf{\mathcal{I}}_{hi})$ using cosine similarity and an $\ell_2$ reconstruction term:
\begin{equation}
\mathcal{L}_{global} = 1 - \cos(\mathbf{z}_{hi}, \hat{\mathbf{z}}_{hi}) + \left\| \mathbf{z}_{hi} - \hat{\mathbf{z}}_{hi} \right\|_2,
\end{equation}
where $\cos(\cdot,\cdot)$ is computed by averaging the cosine similarity over spatial locations.

\noindent {\bf Local Self-supervision.} To enforce fine-grained, spatially consistent reconstruction, we randomly crop a high-resolution view $\mathbf{\mathcal{I}}_{c}\in\mathbb{R}^{M'\times N'\times 3}$ from $\mathbf{\mathcal{I}}_{hi}$ and construct a matched low-resolution view $\mathbf{\mathcal{I}}'\in\mathbb{R}^{M'\times N'\times 3}$ by appropriately downsampling $\mathbf{\mathcal{I}}_{hi}$. Let $\mathbf{z}_c=\phi(\mathbf{\mathcal{I}}_c)$ and $\mathbf{z}'=\phi(\mathbf{\mathcal{I}}')$ be the corresponding features. We upsample $\mathbf{z}'$ to obtain $\mathbf{z}''=\mathcal{G}(\mathbf{z}',\mathbf{\mathcal{I}}')$ and align the crop feature $\mathbf{z}_c$ with the corresponding region $\tilde{\mathbf{z}}$ in $\mathbf{z}''$:
\begin{equation}
    \mathcal{L}_{local} = 1 - \cos(\mathbf{z}_c, \tilde{\mathbf{z}}) + \left\| \mathbf{z}_c - \tilde{\mathbf{z}} \right\|_2.
\end{equation}

Following AnyUp~\cite{anyup}, we further include a self-consistency regularizer $\mathcal{L}_{self}$ to improve robustness of the upsampling process using the same cosine and $\ell_2$ terms.
Finally, the total training objective is:
\begin{equation}
    \mathcal{L} = \mathcal{L}_{global} + \mathcal{L}_{local} + \mathcal{L}_{self}.
\end{equation}

\section{Experiments}
We evaluate our method against FeatUp~\cite{featup}, LiFT~\cite{lift}, LoftUp~\cite{loftup}, JAFAR~\cite{jafar} and AnyUp~\cite{anyup} across a diverse set of downstream tasks, including linear probing semantic segmentation, video object segmentation, object discovery, and keypoint correspondence. \textit{\textbf{Additional results are provided in the Appendix}}.

\subsection{Experimental Settings}
Following prior work~\cite{lift, jafar, anyup}, we train our model on ImageNet dataset for 100K steps using the AdamW optimizer~\cite{adamw}. Unless otherwise specified, we use a batch size of 4 and a learning rate of 1e-3.  For the local part of self-supervised training paradigm, we randomly crop 4 local images for each training image. All experiments are conducted on a single NVIDIA A100 GPU.  By default, we adopt DINOv2-ViT-S/14~\cite{oquab2023dinov2} as the frozen pre-trained VFM.

\begin{table*}[t]
\caption{Performance comparisons on linear probing semantic segmentation on Cityscapes, video object segmentation on DAVIS, and unsupervised object discovery on COCO20K. For semantic segmentation, we report results with default upsampling of each method, while DAVIS and COCO20K are evaluated across different input resolutions using $2\times$ upsampling for all methods because evaluating JAFAR~\cite{jafar} and AnyUP~\cite{anyup} at higher upsampling ratios causes \texttt{OOM} errors on an NVIDIA A100 (80G) GPU for both DAVIS and COCO20K.}
\label{tab: cityscapes vos od}
\vspace{-1.0mm}
\setlength\tabcolsep{4.2pt}
\begin{center}
\begin{small}
\resizebox{\textwidth}{!}{
\begin{tabular}{lcccccccccccc}
    \toprule
    \multirow{4}{*}{\textbf{Method}} 
    & \multicolumn{2}{c}{\textbf{Semantic Segmentation}} 
    & \multicolumn{5}{c}{\textbf{Video Object Segmentation}} 
    & \multicolumn{5}{c}{\textbf{Object Discovery}} \\ 
    \cmidrule(lr){2-3} \cmidrule(lr){4-8} \cmidrule(lr){9-13}
    
    & \multicolumn{2}{c}{Cityscapes} 
    & \multicolumn{5}{c}{$\mathcal{J}\&\mathcal{F}$ Mean ($\uparrow$)} 
    & \multicolumn{5}{c}{CorLoc ($\uparrow$)} \\ 
    \cmidrule(lr){2-3} \cmidrule(lr){4-8} \cmidrule(lr){9-13}

    & mIoU ($\uparrow$) & Acc ($\uparrow$)
    & 448 & 504 & 560 & 672 & 784 
    & 448 & 504 & 560 & 672 & 784 \\
    \cmidrule(rl){1-13}

    DINOv2~\cite{dinov2} 
    & -- & --
    & 54.16 & 58.29 & 61.12 & 65.03 & 67.53 
    & 39.74 & 40.94 & 41.64 & 43.21 & 43.99 \\

    Bilinear 
    & 59.74 & 92.55
    & 53.13 & 55.67 & 59.37 & 62.52 & 64.99 
    & 36.08 & 37.42 & 38.48 & 40.35 & 41.50 \\

    FeatUp~\cite{featup}
    & 60.41 & 93.16 
    & -- & -- & -- & -- & -- 
    & -- & -- & -- & -- & -- \\
    
    LoftUp~\cite{loftup}
    & 61.94 & 93.51 
    & -- & -- & -- & -- & -- 
    & -- & -- & -- & -- & -- \\

    LiFT~\cite{lift} 
    & 60.08 & 92.90
    & 59.25 & 63.06 & 65.76 & 68.86 & 69.78 
    & 39.46 & 40.62 & 41.32 & 42.47 & 43.07 \\

    JAFAR~\cite{jafar}  
    & 61.87 & 93.51
    & 60.37 & 63.06 & 64.01 & 63.59 & 65.05 
    & 25.20 & 25.34 & 25.48 & 25.91 & 26.33 \\

    AnyUp~\cite{anyup}  
    & 61.45 & 93.44
    & 65.55 & 67.17 & 69.03 & 71.21 & 72.44 
    & 26.59 & 27.15 & 27.86 & 28.89 & 30.23 \\

    \cmidrule(rl){1-13}
    Ours  
    & \textbf{62.02} & \textbf{93.51}
    & \textbf{66.09} & \textbf{69.28} & \textbf{71.35} & \textbf{73.57} & \textbf{74.39} 
    & \textbf{41.34} & \textbf{42.52} & \textbf{43.47} & \textbf{44.87} & \textbf{45.65} \\

    \bottomrule
\end{tabular}
}
\end{small}
\end{center}
\vspace{-2.0mm}
\end{table*}

\subsection{Main Results}

\noindent{\bf Linear Probing Semantic Segmentation.} Following the linear probe scheme  adopted in prior works~\cite{jafar, anyup} for semantic segmentation, we train a linear projection head to predict the semantic class labels. Before the head, features extracted by the pre-trained encoder are upsampled to the input resolution by the upsampler. 
As shown in Table~\ref{tab: cityscapes vos od}, our framework achieves competitive results with a lower upsampling factor ($2\times$) on these datasets. Comparing with $14\times$ upsampling methods (JAFAR~\cite{jafar}, and AnyUp~\cite{anyup}), our method achieves superior performance.

\begin{wraptable}{r}{7cm}
\vspace{-4mm}
\caption{Comparison results on the keypoint correspondence task on SPair-71k~\cite{min2019spair}, we report PCK@0.01 metric at multiple input resolutions.}
\label{tab: keypoint}
\vspace{-2mm}
\setlength\tabcolsep{6pt}
\centering
\resizebox{\linewidth}{!}{
\begin{tabular}{lccccc}
    \toprule
    \textbf{Method} & 56 & 112 & 224 & 448 & 560 \\
    \cmidrule(rl){1-6}
    
    DINOv2~\cite{dinov2} & 0.03 & 0.15 & 0.96 & 2.94 & 4.52 \\
    Bilinear & 0.01 & 0.37 & 0.98 & 3.52 & 4.34 \\
    LiFT~\cite{lift} & 0.03 & \textbf{0.50} & 1.46 & 4.32 & 5.60 \\
    JAFAR~\cite{jafar} & 0.02 & 0.43 & 1.30 & 2.58 & 2.64 \\
    AnyUp~\cite{anyup} & 0.01 & 0.42 & 1.28 & 4.05 & 4.50 \\
    \cmidrule(rl){1-6}
    Ours & \textbf{0.07} & 0.47 & \textbf{1.87} & \textbf{5.21} & \textbf{6.51} \\

    \bottomrule
\end{tabular}}
\vspace{-2mm}
\end{wraptable}
\noindent{\bf Video Object Segmentation.} 
We validate our model on video object segmentation, where the ground-truth mask of the first frame is propagated to subsequent frames for mask generation via dense feature matching. Upsampler is used to upsample the features extracted by the pre-trained VFM encoder.
Table~\ref{tab: cityscapes vos od} presents the $\mathcal{J}\&\mathcal{F}$ mean metric on the DAVIS~\cite{davis} dataset across various resolutions. Due to the high computational cost of directly upsampling to the input resolution, all methods perform $2\times$ upsampling. Under this setting, our method demonstrates superior performance compared to existing methods. Notably, our method surpasses AnyUp by a margin of 2.36\% at 672 resolution, highlighting its superior adaptability to high-resolution inputs.

\noindent{\bf Unsupervised Object Discovery.}
We validate our method on unsupervised object discovery. TokenCut~\cite{tokencut} is utilized to perform Graph Cut on features. We report the Correct Localization (CorLoc) metric, which is defined as the fraction of images where at least one predicted bounding box has an IoU greater than 0.5 with the ground truth-box, on COCO20K~\cite{coco} across various resolutions. As shown in Table~\ref{tab: cityscapes vos od}, WRC outperforms existing methods.

\noindent{\bf Keypoint Correspondence.}
We evaluate our method on the keypoint matching task, which involves extracting features via a backbone, upsampling them with an upsampler, and resizing to the input resolution using Lanczos interpolation before performing keypoint matching between image pairs. As shown in Table~\ref{tab: keypoint}, our method consistently outperforms existing works on the PCK@0.01 metric, which demonstrates the effectiveness of our method. Notably, it achieves a 2.27\% improvement over DINOv2 at 448 input resolution.

\begin{figure*}[t]
  \begin{center}
\centerline{\includegraphics[width=\linewidth]{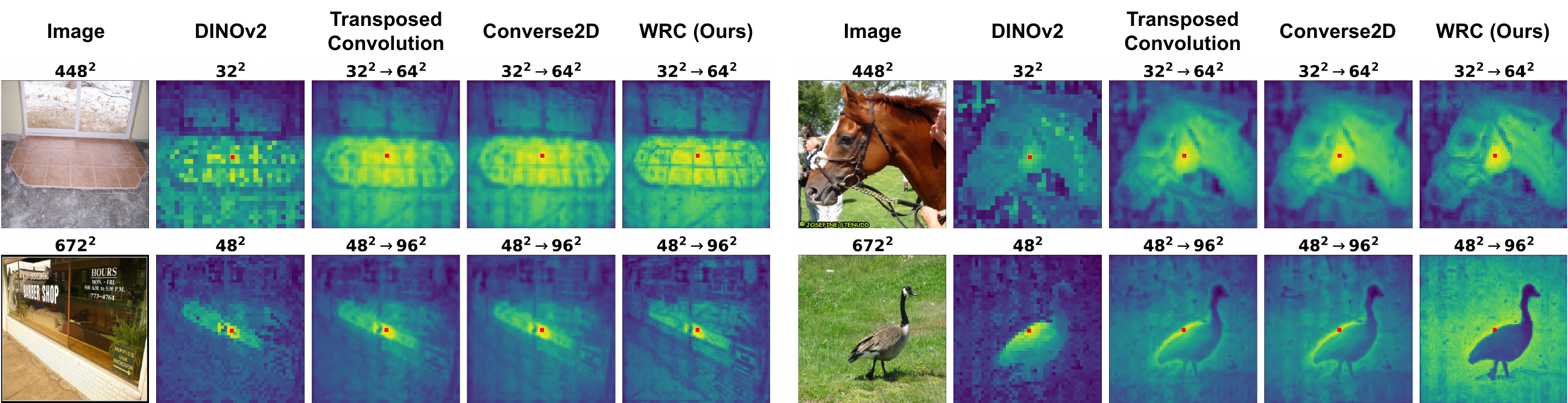}}
    \caption{Qualitative comparison of feature upsampling operators. We visualize similarity maps at two input resolutions, comparing the DINOv2 features with features upsampled by Transposed Convolution, Converse2D, and our WRC. WRC yields sharper, more localized activations around the queried point (\textcolor{red}{red} dot).
    }
    \label{fig:visualization}
    \vspace{-5mm}
  \end{center}
\end{figure*}

\begin{table*}[t]
\setlength\tabcolsep{5pt}
\caption{Comparison of upsampling operators, including Transposed Convolution, Converse2D, and our Weighted Reverse Convolution (WRC), within the same framework for a fair comparison. 
In addition to results on DAVIS and COCO20K, we include open-vocabulary segmentation evaluated with MaskCLIP~\cite{maskclip} using CLIP-ViT-B/16~\cite{clip} as VFM.}
\label{tab: operator ablation}
\begin{center}
\begin{small}
\vspace{-2.5mm}
\begin{tabular}{lcccccccc}
    \toprule
    \multirow{4}{*}{\textbf{Method}} & \multicolumn{3}{c}{\textbf{Video Object Seg.}} & \multicolumn{3}{c}{\textbf{Object Discovery}} & \multicolumn{2}{c}{\textbf{Open-Voc. Seg.}} \\
    \cmidrule(rl){2-4} \cmidrule(rl){5-7} \cmidrule(rl){8-9}
     & \multicolumn{3}{c}{$\mathcal{J}\&\mathcal{F}$ Mean ($\uparrow$)}  & \multicolumn{3}{c}{CorLoc ($\uparrow$)} & \multirow{2}{*}{mIoU ($\uparrow$)} & \multirow{2}{*}{aAcc ($\uparrow$)} \\
    \cmidrule(rl){2-4} \cmidrule(rl){5-7}
     & 448 & 560 & 672 & 448 & 560 & 672 &  &   \\
    \cmidrule(rl){1-9}
    
    Transposed Convolution~\cite{noh2015learning} & 61.35 & 66.73 & 69.68 & 41.04 & 42.51 & 43.70 & 56.98 & 72.99  \\
    Converse2D~\cite{conversenet} & 61.37 & 66.60 & 68.93 & 39.38 & 40.99 & 41.93 & 58.18 & 73.34\\
    WRC (Ours) & \textbf{62.73} & \textbf{68.25} & \textbf{70.71} & \textbf{42.30} & \textbf{44.58} & \textbf{46.13} & \textbf{59.21} & \textbf{73.72} \\
    \bottomrule
\end{tabular}
\end{small}
\end{center}
\vspace{-4mm}
\end{table*}

\subsection{Comparison with Other Upsampling Operators}
\begin{wraptable}{r}{7.8cm}
\vspace{-5mm}
\caption{\textbf{Efficiency and overhead} on DINOv2-ViT-S/14 with $2\times$ feature upsampling and $448\times448$ inputs.
We report the frozen-backbone baseline and the \emph{additional} overhead introduced by each upsampler in parameters and MACs. Inference time is measured with single-image inference on  single NVIDIA A100 GPU. \textbf{Bold} indicates the best result and \underline{underline} indicates the second best.}
\vspace{-2mm}
\label{tab: efficiency}
\setlength{\tabcolsep}{9.8pt}
\centering
\resizebox{\linewidth}{!}{
\begin{tabular}{llll}
\toprule
\textbf{Method} & {Param (M)} & {MACs (G)} & \textbf{Time (ms)}  \\
\midrule
DINOv2~\cite{dinov2}& 22.06 & 31.79  & 5.99   \\
\midrule
FeatUp~\cite{featup}  & +0.17 & +51.12   & 65.13  \\
LoftUp~\cite{loftup}  & +4.30 & +1066.20 & 134.73  \\
JAFAR~\cite{jafar}   & +0.63 & +51.35   & 59.35   \\
AnyUp~\cite{anyup}   & +0.88 & +50.79   & 61.55   \\
LiFT~\cite{lift}    & +1.19 & +3.90    & \textbf{6.79}   \\
\textbf{Ours} & +3.98 & +10.34 & \underline{9.45}   \\
\bottomrule
\end{tabular}}
\vspace{-5mm}
\end{wraptable}
Table~\ref{tab: operator ablation} shows the comparison results between our WRC and other upsampling operators.
We keep the same setting for different methods to make a fair comparison.
In the open-vocabulary semantic segmentation task, pixel-wise predictions are generated by selecting the class label with the highest similarity score between the textual class embeddings and the image features as in ~\cite{maskclip}.  
We utilize the CLIP-ViT-B/16 backbone and upsample its output features using our proposed method.  
Our WRC outperforms the Transposed Convolution~\cite{noh2015learning} by +2.43\% on the object discovery task at the resolution of 672. Fig.~\ref{fig:visualization} further provides qualitative comparisons on similarity maps obtained from 
features at different input resolutions.

\subsection{Efficiency and Overhead}
Table~\ref{tab: efficiency} reports the efficiency of different feature upsampling approaches in terms of parameter count, computational cost (MACs), and wall-clock inference latency. For a fair comparison, all methods use the same 2$\times$ upsampling setting and the same  448$\times$448 input resolution, and all timing results are measured on the same single NVIDIA A100 GPU. Compared with universal high-capacity upsamplers (\textit{e.g.}, FeatUp, JAFAR, AnyUp, and especially LoftUp), WRC introduces substantially lower computational overhead and runs markedly faster at inference, while delivering stronger dense features in downstream evaluations. At the same time, compared to LiFT, our approach remains highly efficient yet consistently improves performance across tasks. These results indicate that incorporating a closed-form inverse operator with lightweight learned weighting can provide practical gains without sacrificing  efficiency.

\subsection{Ablation Study}

\begin{wraptable}{r}{7.5cm}
\vspace{-4mm}
\caption{\textbf{Implementation ablations} of WRC.  
We evaluate weight initialization and positive weight parameterization across multiple input resolutions. 
}
\label{tab:implementation}
\setlength{\tabcolsep}{4pt}
\centering
\resizebox{\linewidth}{!}{
\begin{tabular}{lcccccc}
\toprule
\textbf{Setting} &
\multicolumn{3}{c}{\textbf{Video Object Seg.}} &
\multicolumn{3}{c}{\textbf{Object Discovery}} \\
\cmidrule(lr){2-4}\cmidrule(lr){5-7}
& \multicolumn{3}{c}{$\mathcal{J}\&\mathcal{F}$ Mean ($\uparrow$)}
& \multicolumn{3}{c}{CorLoc ($\uparrow$)} \\
\cmidrule(lr){2-4}\cmidrule(lr){5-7}
& 448 & 560 & 672 & 448 & 560 & 672 \\
\midrule

\multicolumn{7}{l}{\textbf{\emph{Weight Initialization}}} \\
\hspace{0.6em}Matrix
& \textbf{65.77} & 69.95 & 72.42
& 40.61 & 42.56 & 44.02 \\
\hspace{0.6em}Convolution
& 65.71 & \textbf{70.68} & \textbf{73.07}
& \textbf{40.94} & \textbf{42.78} & \textbf{44.13} \\

\addlinespace[2pt]
\multicolumn{7}{l}{\textbf{\emph{Weight Parameterization}}} \\
\hspace{0.6em}None
& 65.71 & 70.68 & 73.07
& 40.94 & 42.78 & 44.13 \\
\hspace{0.6em}Softplus
& 65.66 & 70.40 & 72.65
& 40.66 & 42.51 & 43.53 \\
\hspace{0.6em}Log1p 
& \textbf{66.09} & \textbf{71.35} & \textbf{73.57}
& \textbf{41.34} & \textbf{43.47} & \textbf{44.87} \\
\bottomrule
\end{tabular}}
\vspace{-2mm}
\end{wraptable}
We conduct ablations using DINOv2-ViT-S/14 on DAVIS video object segmentation and COCO20K unsupervised object discovery, evaluated across multiple input resolutions.

\noindent{\bf Weight Initialization and Parameterization.}
We analyze key implementation choices of WRC on two representative dense benchmarks. As discussed in Sec.~\ref{sec: implementation of WRC}, we focus on two factors: how the spatial weight fields are generated (\textit{initialization}) and how positivity is enforced (\textit{parameterization}).
We compare matrix-based, input-agnostic weights with convolution-based, input-adaptive weight predictors. Convolution-based initialization consistently improves performance on both tasks, yielding higher $\mathcal{J}\&\mathcal{F}$ at 560/672 and better CorLoc at all resolutions.
Besides, we evaluate different positivity schemes for the weights: none, softplus, and log1p. Log1p achieves the best and most consistent results across both tasks and input resolutions, outperforming the unregularized variant and softplus.

\begin{wraptable}{r}{9.0cm}
\vspace{-4mm}
\caption{\textbf{Training objective ablations} of WRC.  
We evaluate three training objectives 
across input resolutions. 
\textbf{Bold} indicates the best result and \underline{underline} indicates the second best.}
\vspace{-1mm}
\label{tab:loss}
\setlength{\tabcolsep}{7pt}
\centering
\resizebox{\linewidth}{!}{
\begin{tabular}{ccccccc}
\toprule
\multicolumn{3}{l}{\textbf{Setting}} & \multicolumn{2}{c}{\textbf{Video Object Seg.}} & \multicolumn{2}{c}{\textbf{Object Discovery}} \\
\cmidrule(lr){4-5}\cmidrule(lr){6-7}
 &  &  & \multicolumn{2}{c}{$\mathcal{J}\&\mathcal{F}$ Mean ($\uparrow$)} & \multicolumn{2}{c}{CorLoc ($\uparrow$)} \\
\cmidrule(lr){4-5}\cmidrule(lr){6-7}
$\mathcal{L}_{global}$ & $\mathcal{L}_{local}$ & $\mathcal{L}_{self}$ & 448 & 672 & 448 & 672 \\
\midrule
\checkmark &  &  & 63.28 & 71.22 & \underline{41.19} & \textbf{44.88}\\
 & \checkmark &  & \textbf{66.64} & \textbf{73.69} & 39.37 & 42.57\\
 &  & \checkmark &  20.63 & 26.31 & 1.27 & 0.98\\
\checkmark & \checkmark &  & 65.81 & 73.05 & 40.21 & 43.65\\
\checkmark &  & \checkmark & 62.85 & 70.58 & 41.05 & 44.44\\
\checkmark & \checkmark & \checkmark & \underline{66.09} & \underline{73.57} & \textbf{41.34} & \underline{44.87}\\
\bottomrule
\end{tabular}}
\vspace{-6mm}
\end{wraptable}
\noindent{\bf Effects of Training Objectives.}
Table~\ref{tab:loss} ablates the three loss terms $\mathcal{L}_{global}$, $\mathcal{L}_{local}$, and $\mathcal{L}_{self}$. $\mathcal{L}_{global}$ alone yields strong object discovery but weaker segmentation, whereas $\mathcal{L}_{local}$ alone improves segmentation yet hurts object discovery, showing the two signals are complementary. $\mathcal{L}_{self}$ by itself performs poorly, indicating it mainly acts as a regularizer. Combining all three losses produces the most balanced and consistent results, achieving the best (or tied-best) CorLoc while maintaining near-top segmentation performance across resolutions.

\subsection{Qualitative Analysis}

Figure~\ref{fig:visualization} visualizes similarity maps obtained from DINOv2 features at different input resolutions.
We compare the original coarse descriptors with features densified by transposed convolution, Converse2D, and our WRC.
Across all examples, the raw DINOv2  maps exhibit blocky activations and limited spatial precision. In contrast, our approach consistently produces fine-grained, high-contrast peaks centered on the queried point, with clearer object extent.
Besides, we further provide qualitative results for linear probing semantic segmentation in Figure~\ref{fig:visualization2} in \textbf{Appendix}. \textbf{\textit{The section~\ref{sec} in Appendix provides more qualitative results}}.
Overall, these visualizations show: the proposed adaptive, weighted inversion better preserves correspondence-critical structures while maintaining stable feature distributions.

\section{Conclusion and Limitations}
This paper revisited feature upsampling for vision foundation models through an inverse-problem formulation.  
We proposed Weighted Reverse Convolution (WRC), a principled inverse operator that generalizes reverse convolution with spatially adaptive data fidelity and Tikhonov regularization via positive, learned per-location weights. WRC admitted an efficient, fully differentiable closed-form FFT solution, making it a practical drop-in replacement for conventional upsampling blocks. Across a broad range of downstream benchmarks, including linear probing segmentation, video object segmentation, object discovery, and keypoint correspondence, our method consistently improved dense feature quality while maintaining high efficiency.

\textbf{Limitations}. Although WRC achieves competitive inference efficiency among recent feature upsampling methods, it remains slightly less efficient than very simple operators such as transposed convolution, due to its FFT-based solving and adaptive weight prediction. Further lightweight optimization remains an important direction for future work.


{\small
\bibliography{ref}
}

\appendix

\clearpage
\setcounter{table}{0}
\setcounter{equation}{0}
\setcounter{figure}{0}
\renewcommand{\thetable}{\thesection.\arabic{table}}
\renewcommand{\theequation}{\thesection.\arabic{equation}}
\renewcommand{\thefigure}{\thesection.\arabic{figure}}
\begin{center}
    {\LARGE\bfseries Appendix}
\end{center}

In this appendix, we provide the following materials:
\begin{itemize}
    \item \textbf{A.} BCCB Patterns in DINOv3
    \item \textbf{B.} Proof of Closed-form Solution for WRC
    \item \textbf{C.} Additional PyTorch-like Implementation Details
    \item \textbf{D.} More Experimental Results
\end{itemize}

\section{BCCB Patterns in DINOv3}\label{sec:bccb}

\begin{figure*}[h]
  \vspace{-2mm}
  \begin{center}
    \centerline{\includegraphics[width=1\linewidth]{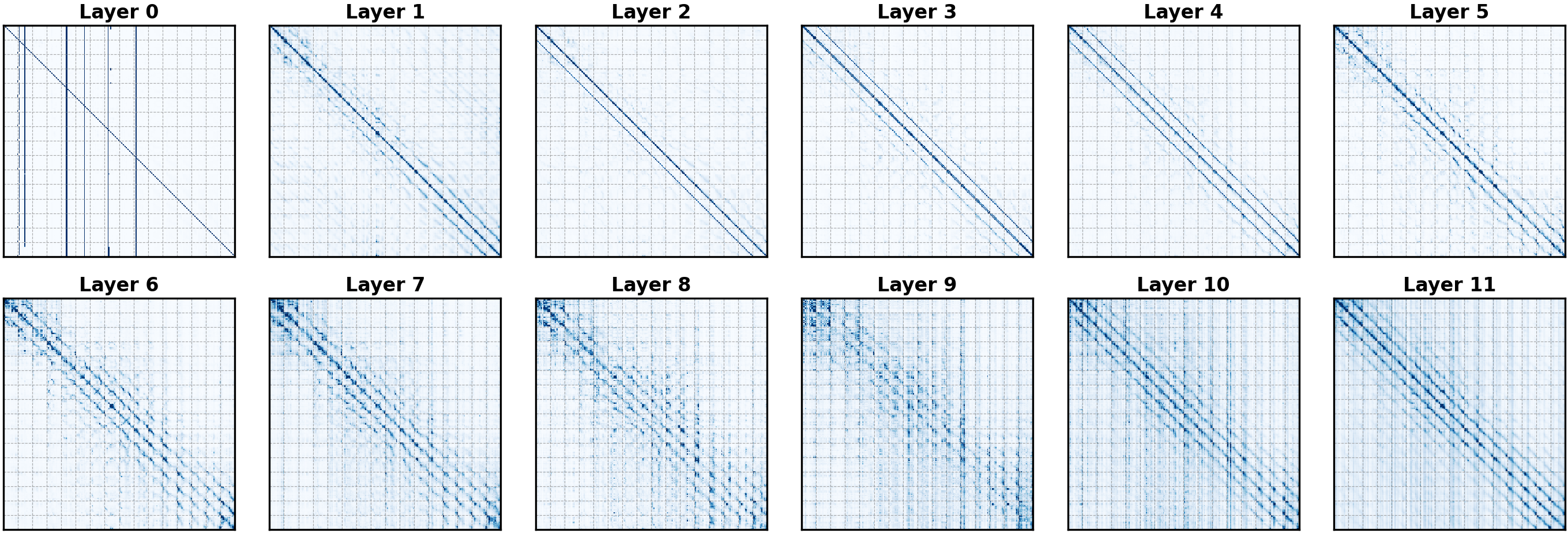}}
    \caption{Per-layer attention maps averaged over all heads in DINOv3~\cite{simeoni2025dinov3}. Without handcrafted constraints, the self-attention module in Vision Transformer learns near BCCB patterns.}
    \label{fig: appendix attention dinov3}
    \vspace{-5.8mm}
  \end{center}
\end{figure*}

We also observe near BCCB patterns in DINOv3, as shown in Fig.~\ref{fig: appendix attention dinov3}. Its patch-to-patch attention maps often exhibit clear block-wise circulant structures, suggesting that such convolution-like behavior is not unique to DINOv2 but commonly appears across different VFM backbones.
This provides broader motivation to apply reverse convolution to VFM features: the backbone's attention itself already behaves in a translation-sharing and Fourier-friendly manner.

\section{Proof of Closed-form Solution for WRC}\label{sec:proof}
\subsection{General Case: $s>1$}
Consider the following optimization problem:
\begin{align}
    \min_{\mathbf{x}} \phi(\mathbf{x}) = \left\| \mathbf{W} \left( \mathbf{y} - \mathbf{S}\mathbf{K}\mathbf{x} \right) \right\|_2^2 + \left\| \mathbf{W}_{\lambda} \left( \mathbf{F}\mathbf{x} - \mathbf{F}\mathbf{x}_0 \right) \right\|_2^2,
\end{align}
where:
\begin{itemize}
    \item $N_l = h \times w$ and $N_h = H \times W$ denote the dimensions, with the scaling relation $N_h = N_l \cdot s^2$ (where $s$ is the scale factor).
    \item $\mathbf{y} \in \mathbb{R}^{N_l}$ is the observed low-resolution feature, and $\mathbf{x} \in \mathbb{R}^{N_h}$ is the high-resolution feature to be estimated.
    \item $\mathbf{x}_{0} \in \mathbb{R}^{N_h}$ is the prior reference feature.
    \item $\mathbf{S} \in \mathbb{R}^{N_l \times N_h}$ represents the down-sampling matrix.
    \item $\mathbf{K} \in \mathbb{R}^{N_h \times N_h}$ is the convolution kernel matrix.
    \item $\mathbf{F} \in \mathbb{C}^{N_h \times N_h}$ denotes the unitary Fourier transform matrix.
    \item $\mathbf{W} \in \mathbb{R}^{N_l \times N_l}$ and $\mathbf{W}_{\lambda} \in \mathbb{R}^{N_h \times N_h}$ are positive-definite diagonal weight matrices.
\end{itemize}

Exploiting the property that the circular convolution matrix $\mathbf{K}$ is diagonalizable by the Fourier transform, we have:
\begin{subequations}
\begin{align}
    \mathbf{K} &= \mathbf{F}^H \bm{\Lambda} \mathbf{F}, \\
    \mathbf{K}^H &= \mathbf{F}^H \bm{\Lambda}^H \mathbf{F},
\end{align}
\end{subequations}
where $\bm{\Lambda} = \mathrm{diag}\left( \mathbf{Fk} \right) \in \mathbb{R}^{N_h \times N_h}$.

Setting the derivative of $\phi(\mathbf{x})$ with respect to $\mathbf{x}$ to zero yields:
\begin{subequations}
\begin{align}
    \frac{\partial \phi}{\partial \mathbf{x}} &= -2\mathbf{K}^H\mathbf{S}^H\mathbf{W}^H (\mathbf{W}\mathbf{y}-\mathbf{W}\mathbf{S}\mathbf{K}\mathbf{x}) + 2\mathbf{F}^H\mathbf{W}_\lambda^H (\mathbf{W}_\lambda \mathbf{F}\mathbf{x} - \mathbf{W}_\lambda \mathbf{F} \mathbf{x}_0) \\
    &= -2 (\mathbf{K}^H \mathbf{S}^H \mathbf{W}^H \mathbf{W} \mathbf{y} + \mathbf{F}^H \mathbf{W}_\lambda^H \mathbf{W}_\lambda \mathbf{F} \mathbf{x}_0) + 2 (\mathbf{K}^H \mathbf{S}^H \mathbf{W}^H \mathbf{W} \mathbf{S} \mathbf{K} \mathbf{x} + \mathbf{F}^H \mathbf{W}_\lambda^H \mathbf{W}_\lambda \mathbf{F} \mathbf{x})\\
    &= -2 (\mathbf{K}^H \mathbf{S}^H \lvert \mathbf{W}\rvert^{2} \mathbf{y} + \mathbf{F}^H \lvert \mathbf{W}_\lambda\rvert^{2} \mathbf{F} \mathbf{x}_0) + 2 (\mathbf{K}^H \mathbf{S}^H \lvert \mathbf{W}\rvert^{2} \mathbf{S} \mathbf{K} \mathbf{x} + \mathbf{F}^H \lvert \mathbf{W}_\lambda\rvert^{2} \mathbf{F} \mathbf{x})\\
    &= \mathbf{0}.
\end{align}
\end{subequations}

Rearranging the terms, we obtain the normal equation:
\begin{subequations}
\begin{align}
    \mathbf{K}^H \mathbf{S}^H \lvert \mathbf{W} \rvert^2 \mathbf{S}\mathbf{K}\mathbf{x} + \mathbf{F}^H \lvert \mathbf{W}_\lambda \rvert^2 \mathbf{F}\mathbf{x} 
     &= \mathbf{K}^{H}\mathbf{S}^{H}\lvert \mathbf{W}\rvert^{2} \mathbf{y} + \mathbf{F}^{H}\lvert \mathbf{W}_{\lambda}\rvert^{2} \mathbf{F} \mathbf{x}_{0} \\
     &= \left(\mathbf{F}^{H}\bm{\Lambda}^{H}\mathbf{F}\right) \mathbf{S}^{H}\lvert \mathbf{W}\rvert^{2} \mathbf{y} + \mathbf{F}^{H}\lvert \mathbf{W}_{\lambda}\rvert^{2} \mathbf{F} \mathbf{x}_{0} \\
     &= \mathbf{F}^{H} \left( \bm{\Lambda}^{H}\mathbf{F} \mathbf{S}^{H}\lvert \mathbf{W}\rvert^{2} \mathbf{y} + \lvert \mathbf{W}_{\lambda}\rvert^{2} \mathbf{F} \mathbf{x}_{0} \right).
\end{align}
\end{subequations}
Let $\mathbf{G}=\bm{\Lambda}^{H}\mathbf{F} \mathbf{S}^{H}\lvert \mathbf{W}\rvert^{2} \mathbf{y} + \lvert \mathbf{W}_{\lambda}\rvert^{2} \mathbf{F} \mathbf{x}_{0}$. 
Solving for $\mathbf{x}$, we derive:
\begin{subequations}
\begin{align}
    \mathbf{x} &= \left( \mathbf{K}^H \mathbf{S}^H \lvert \mathbf{W} \rvert^2 \mathbf{SK} + \mathbf{F}^H \lvert \mathbf{W}_{\lambda} \rvert^2 \mathbf{F} \right)^{-1} \mathbf{F}^H \mathbf{G}, \\
    &= \mathbf{F}^H \left( \mathbf{F}\mathbf{K}^H \mathbf{S}^H \lvert \mathbf{W} \rvert^2 \mathbf{S}\mathbf{K}\mathbf{F}^H + \lvert \mathbf{W}_{\lambda} \rvert^2 \right)^{-1}\mathbf{G} \\
    &= \mathbf{F}^H \left[ \mathbf{F}\mathbf{K}^H \mathbf{F}^H \left(\mathbf{F} \mathbf{S}^H \lvert \mathbf{W} \rvert^2 \mathbf{S} \mathbf{F}^H\right) \mathbf{F} \mathbf{K}\mathbf{F}^H + \lvert \mathbf{W}_{\lambda} \rvert^2 \right]^{-1}\mathbf{G} \\
    &= \mathbf{F}^H \left[ \bm{\Lambda}^H \left(\mathbf{F} \mathbf{S}^H \lvert \mathbf{W} \rvert^2 \mathbf{S} \mathbf{F}^H\right) \bm{\Lambda} + \lvert \mathbf{W}_{\lambda} \rvert^2 \right]^{-1}\mathbf{G} \label{eq: prof_solution_1}.
\end{align}
\end{subequations}

Let $\lvert \mathbf{W} \rvert^2 = \mathrm{diag}(\lvert W_1 \rvert^2, \ldots, \lvert W_{N_l} \rvert^2) \in \mathbb{R}^{N_l \times N_l}$. The up-sampled weight matrix in the spatial domain is: 
\begin{equation}
    \mathbf{S}^H \lvert \mathbf{W} \rvert^2 \mathbf{S} = \mathrm{diag}\bigl(\lvert W_1 \rvert^2, 0, \ldots, 0, \lvert W_2 \rvert^2, 0, \ldots, 0, \lvert W_{N_l} \rvert^2, 0, \dots, 0 \bigr) \in \mathbb{R}^{N_h\times N_h}.
\end{equation}
Then, we have:
\begin{equation}
    \mathbf{F}\mathbf{S}^H \lvert \mathbf{W} \rvert^2 \mathbf{S}\mathbf{F}^H = \frac{1}{d} \mathbf{J}_d \otimes \left( \mathbf{F}\lvert \mathbf{W} \rvert^2 \mathbf{F}^H \right),
\end{equation}
where $d = s^2$, $\mathbf{J}_d \in \mathbb{R}^{d \times d}$ is a matrix of ones, and $\otimes$ denotes the Kronecker product. Substituting this into the bracketed term of Eq.~\eqref{eq: prof_solution_1}:
\begin{subequations}
\begin{align}
    \bm{\Lambda}^H \left(\mathbf{F} \mathbf{S}^H \lvert \mathbf{W} \rvert^2 \mathbf{S} \mathbf{F}^H\right) \bm{\Lambda}
    &= \mathrm{diag}\left(\bm{\Lambda}_1^H, \ldots, \bm{\Lambda}_d^H \right) \left[\frac{1}{d}\mathbf{J}_d \otimes \left( \mathbf{F}\lvert \mathbf{W} \rvert^2 \mathbf{F}^H \right)\right] \mathrm{diag}\left(\bm{\Lambda}_1, \ldots, \bm{\Lambda}_d \right) \\
    &= \frac{1}{d} \begin{bmatrix} \bm{\Lambda}_1^H \\ \vdots \\ \bm{\Lambda}_d^H \end{bmatrix} \left( \mathbf{F}\lvert \mathbf{W} \rvert^2 \mathbf{F}^H \right) \begin{bmatrix} \bm{\Lambda}_1 & \cdots & \bm{\Lambda}_d \end{bmatrix},
\end{align}
\end{subequations}
where $\bm{\Lambda}_i \in \mathbb{C}^{N_l \times N_l}$ represents the $i$-th sub-block of the diagonal kernel matrix $\bm{\Lambda}$.

Let $\bm{\underline{\Lambda}} = \begin{bmatrix} \bm{\Lambda}_1 & \cdots & \bm{\Lambda}_d \end{bmatrix} \in \mathbb{C}^{N_l \times N_h}$ and let $\mathbf{C} = \frac{1}{d} \mathbf{F}\lvert \mathbf{W} \rvert^2 \mathbf{F}^H$. Eq.~\eqref{eq: prof_solution_1} can be reformulated as:
\begin{equation}
    \mathbf{x} = \mathbf{F}^H \left[ \lvert \mathbf{W}_{\lambda} \rvert^2 + \bm{\underline{\Lambda}}^H \mathbf{C} \bm{\underline{\Lambda}} \right]^{-1}\mathbf{G}.
    \label{eq: prof_solution_2}
\end{equation}

Applying the Woodbury matrix identity (Sherman-Morrison-Woodbury formula), Eq.~\eqref{eq: prof_solution_2} becomes:
\begin{equation}
    \mathbf{x} = \mathbf{F}^H \left[ \mathbf{D}^{-1} - \mathbf{D}^{-1} \bm{\underline{\Lambda}}^H \left( \mathbf{C}^{-1} + \bm{\underline{\Lambda}} \mathbf{D}^{-1} \bm{\underline{\Lambda}}^H \right)^{-1} \bm{\underline{\Lambda}} \mathbf{D}^{-1}  \right] \mathbf{G},
    \label{eq: prof_solution_3}
\end{equation}
where $\mathbf{D} = \lvert \mathbf{W}_{\lambda} \rvert^2$.

Finally, the matrix representation of Eq.~\eqref{eq: prof_solution_3} is:
\begin{equation}
    \mathbf{X} = \mathcal{F}^{-1} \left(\frac{1}{\lvert \mathbf{W}_{\lambda} \rvert^2} \left(\mathbf{L'} - \overline{\mathcal{F}_\mathbf{K}} \odot_s \frac{ \lvert \mathbf{W} \rvert^2 \odot \left( \left( \mathcal{F}_\mathbf{K} \odot \mathbf{L'} \right) \Downarrow_s \right) }{\lvert \mathbf{W} \rvert^2 \odot \left(\lvert \mathcal{F}_\mathbf{K} \rvert^2 \Downarrow_s \right)  + \lvert \mathbf{W}_{\lambda} \rvert^2 \Downarrow_s } \right)\right),
    \label{eq: prof_solution_4}
\end{equation}
where:
\begin{itemize}
    \item $\mathbf{L'} = \overline{\mathcal{F}_\mathbf{K}}  \odot \mathcal{F}_{(\lvert\mathbf{W}\rvert^{2} \odot \mathbf{Y})\uparrow_s} + \lvert \mathbf{W}_{\lambda}\rvert^{2} \odot \mathcal{F}_{\mathbf{X}_0}$.
    \item $\odot$ denotes element-wise multiplication.
    \item $\odot_s$ and $\Downarrow_s$ denote element-wise multiplication applied to $ s\times s$ distinct blocks and the distinct block down-sampling operator that averages over these $ s\times s$ blocks.
\end{itemize}

\subsection{Special Case: $s=1$}

When $s=1$, the problem simplifies to:
\begin{equation}
    \min_{\mathbf{X}} \phi(\mathcal{F}_\mathbf{X}) = \left\| \mathbf{W} \odot \left( \mathcal{F}_\mathbf{Y} - \mathcal{F}_\mathbf{K} \odot \mathcal{F}_\mathbf{X} \right) \right\|_F^2 + \left\| \mathbf{W}_{\lambda} \odot \left( \mathcal{F}_\mathbf{X} - \mathcal{F}_\mathbf{X_0} \right) \right\|_F^2.
\end{equation}
The closed-form solution is derived directly as:
\begin{equation}
    \frac{\partial \phi}{\partial \mathcal{F}_\mathbf{X}} = -2 \lvert \mathbf{W} \rvert^{2} \odot \left( \overline{\mathcal{F}_\mathbf{K}} \odot \mathcal{F}_\mathbf{Y} - \lvert \mathcal{F}_\mathbf{K} \rvert^{2} \odot \mathcal{F}_\mathbf{X} \right) + \lvert \mathbf{W}_\lambda \rvert^{2} \odot \left( \mathcal{F}_\mathbf{X} - \mathcal{F}_\mathbf{X_0} \right) = 0,
\end{equation}
Thus, the estimate is:
\begin{align}
    \mathbf{X} = \mathcal{F}^{-1}\left( \frac{\lvert \mathbf{W} \rvert^{2} \odot \overline{\mathcal{F}_\mathbf{K}} \odot \mathcal{F}_\mathbf{Y} + \lvert \mathbf{W}_\lambda \rvert^{2} \odot \mathcal{F}_\mathbf{X_0} }{ \lvert \mathbf{W} \rvert^{2} \odot \lvert \mathcal{F}_\mathbf{K} \rvert^{2} + \lvert \mathbf{W}_\lambda \rvert^{2}} \right).
\end{align}

\subsection{Degenerate Case: ConverseNet}
By setting $\mathbf{W} = \mathbf{I}_{N_l}$, $\mathbf{C} = \mathbf{I}_{N_l}$, and $\mathbf{W}_\lambda = \sqrt{\lambda} \mathbf{I}_{N_h}$, the proposed solution reduces to the formulation presented in~\cite{conversenet}. In this case, Eq.~\eqref{eq: prof_solution_3} simplifies to:
\begin{align}
    \mathbf{x} &= \mathbf{F}^H \left[ \mathbf{I} - \bm{\underline{\Lambda}}^H \left(\lambda \mathbf{I} + \bm{\underline{\Lambda}} \bm{\underline{\Lambda}}^H \right)^{-1} \bm{\underline{\Lambda}} \right] \mathbf{G'},
\end{align}
where $\mathbf{G'} = \bm{\Lambda}^{H}\mathbf{F} \mathbf{S}^{H} \mathbf{y} + \lambda \mathbf{F} \mathbf{x}_{0}$.
Consequently, its matrix representation can be formulated as:
\begin{align}
    \mathbf{X}&=\mathcal{F}^{-1}\left(\frac{1}{\lambda}{\left(\mathbf{L} - \overline{\mathcal{F}_\mathbf{K}} \odot_s \frac{(\mathcal{F}_\mathbf{K} \odot \mathbf{L})\Downarrow_s}{| \mathcal{F}_\mathbf{K} |^2\Downarrow_s+\lambda}\right)}\right),
\end{align}
where $\mathbf{L} = \overline{\mathcal{F}_\mathbf{K}} \mathcal{F}_{\mathbf{Y}\uparrow_s} + \lambda \mathcal{F}_{\mathbf{X}_0}$.

For the special case of $s = 1$, the equation further simplifies to:
\begin{align}
    \mathbf{X} = \mathcal{F}^{-1}\left( \frac{ \overline{\mathcal{F}_\mathbf{K}} \odot \mathcal{F}_\mathbf{Y} + \lambda \mathcal{F}_\mathbf{X_0} }{ \lvert \mathcal{F}_\mathbf{K} \rvert^{2} + \lambda } \right).
\end{align}

\begin{algorithm}[t]
\caption{Code of WRC in a PyTorch-like style.}
\label{alg: wrc}
\lstset{
    backgroundcolor=\color{white},
    basicstyle=\fontsize{7.4pt}{7.4pt}\ttfamily\selectfont,
    columns=fullflexible,
    breaklines=true,
    captionpos=b,
    commentstyle=\fontsize{8pt}{8pt}\color{cyan},
    keywordstyle=\fontsize{8.0pt}{8.0pt},
}
\begin{lstlisting}[language=python]
# X: input feature map
# K: kernel
# S: stride or upscaling factor
# p2o: convert point-spread function (PSF) to optical transfer function (OTF)
# splits: divide a tensor of shape [..., H, W] into [..., H/S, W/S, S^2] distinct blocks
# upsample: s-fold upsampling operator that inserts zeros
# conv_W, conv_W_lam: convolutions for obtaining W and W_lam

B, C, H, W = X.shape        # X: B, C, H, W
biaseps = nn.functional.softplus(bias) + 1e-5
# W: regularization parameter
W = torch.log1p(torch.pow(conv_W(X), 2))
# upsample X: B, C, HxS, WxS
Y_S = upsample(W*X, scale=S)
# interpolate X: B, C, HxS, WxS
X_0 = nn.functional.interpolate(X, scale_factor=S, mode='bilinear', align_corners=True)
# W_lam: regularization parameter
W_lam = torch.log1p(torch.pow(conv_W_lam(X_0), 2)) 

# FFT kernel: B, C, HxS, WxS
FK = p2o(K, (H*S, W*S))
# complex conjugate
FK_conj = torch.conj(FK)
# element-wise squared magnitude
FK_2 = torch.pow(torch.abs(FK), 2)
FKY = FK_conj*torch.fft.fftn(Y_S, dim=(-2, -1))
# L: B, C, HxS, WxS
L = FKY + torch.fft.fftn((W_lam+biaseps)*X_0, dim=(-2, -1))
FKL = FK.mul(L)
# split and calculate mean: B, C, H, W
FKL_S = W * torch.mean(splits(FKL, S), dim=-1, keepdim=False)
FK2_S = W * torch.mean(splits(FK_2, S), dim=-1, keepdim=False)
Fdiv = FKL_S.div(FK2_S + torch.mean(splits(W_lam, S), dim=-1, keepdim=False) + biaseps)
# element-wise multiplication
Fmul = FK_conj*Fdiv.repeat(1, 1, S, S)
Fout = (L-Fmul)/(W_lam+biaseps)
# inverse FFT output: B, C, HxS, WxS
out = torch.real(torch.fft.ifftn(Fout, dim=(-2, -1)))

return out
\end{lstlisting}
\end{algorithm}

\section{Additional PyTorch-like Implementation Details}\label{sec:algo} 
Algorithm~\ref{alg: wrc} summarizes the practical implementation of our WRC in a PyTorch-like form. Given an input feature map $X$, we first predict two positive, spatially varying weight maps: $W$ for the data-fidelity term (from $X$) and $W_\lambda$ for the regularization term (from the interpolated prior $X_0$). Positivity and stability are enforced via a $\log(1+x)$ re-parameterization.

We then compute the closed-form solution in the Fourier domain: the kernel $K$ is converted to its frequency response, and the main operations are batched FFT/IFFT and element-wise complex arithmetic. The distinct-block averaging required by the $\Downarrow_s$ operator is implemented via \texttt{splits}. A small $\epsilon$ is added to denominators to prevent numerical issues, and the inverse FFT yields the final high-resolution output.

\section{More Experimental Results}\label{sec:experiment}

\subsection{More Results on DINOv3}
Table~\ref{tab:vos dinov3} reports DAVIS video object segmentation results using DINOv3-ViT-S/16~\cite{simeoni2025dinov3} features across input resolutions. DINOv3 adopts a larger patch size, which yields fewer tokens and a coarser feature grid at a fixed input solution, making high-resolution densification more challenging.  Under this setting, prior upsampling methods (JAFAR, AnyUp, and LiFT) provide gains mainly at lower resolutions, while their performance saturates or drops as the input resolution increases (\textit{e.g.}, 672/784), suggesting limited robustness when scaling to finer spatial details. In contrast, our method consistently improves with resolution and delivers the strongest results at high resolutions, achieving clear gains at 560/672/784. This trend indicates that WRC better preserves discriminative structures and remains stable when upsampling dense foundation-model features at higher input resolutions, highlighting its effectiveness and scalability.

\begin{table*}[t]
\caption{DAVIS Video Object Segmentation results using \textbf{DINOv3-ViT-S/16}~\cite{simeoni2025dinov3} reported with the $\mathcal{J}\&\mathcal{F}$ mean, evaluated across input resolutions. Note that evaluating JAFAR~\cite{jafar} and AnyUP~\cite{anyup} at higher upsampling ratios triggers \texttt{OOM} errors with an  NVIDIA A100 (80G); therefore, we use $2\times$ upsampling for all methods.}
\label{tab:vos dinov3}
\setlength\tabcolsep{6.5pt}
\begin{center}
\begin{small}
\begin{tabular}{lcccc}
    \toprule
    \textbf{Method} & 448 & 560 & 672 & 784\\
     \cmidrule(rl){1-5}
    DINOv3~\cite{simeoni2025dinov3} & 56.69 & 63.79 & 67.89 & 69.64\\
    Bilinear & 50.35 & 54.92 & 59.30 & 62.44\\
    LiFT~\cite{lift} & 57.31 & 61.13 & 66.00 & 68.24\\
    JAFAR~\cite{jafar}  & \textbf{61.31} & 64.61 & 67.12 & 67.91\\
    AnyUp~\cite{anyup}  &  57.57 & 60.45 & 62.43 & 64.20\\
    \cmidrule(rl){1-5}
    Ours  &  61.30 & \textbf{65.17} & \textbf{69.10} & \textbf{70.77}\\
    \bottomrule
\end{tabular}
\end{small}
\end{center}
\end{table*}

\begin{table*}[t!]
\caption{COCO2017 results for detection and instance segmentation with Mask R-CNN~\cite{he2017mask}. We replace the default feature upsampling operator in FPN with Converse2D and our WRC, respectively, while keeping the remaining architecture and training protocol unchanged.}
\label{tab:Mask-RCNN}
\begin{center}
\setlength{\tabcolsep}{3pt}
\resizebox{\linewidth}{!}{
\begin{tabular}{lcccccccccccc}
    \toprule
    \multirow{2}{*}{\textbf{Mask R-CNN}} & \multicolumn{6}{c}{\textbf{Detection}} & \multicolumn{6}{c}{\textbf{Segmentation}} \\
    \cmidrule(rl){2-7} \cmidrule(rl){8-13}
    & \text{mAP} & $\text{mAP}_{50}$ & $\text{mAP}_{75}$ & $\text{mAP}_{s}$ & $\text{mAP}_{m}$ & $\text{mAP}_{l}$ & $\text{mAP}$ & $\text{mAP}_{50}$ & $\text{mAP}_{75}$ & $\text{mAP}_{s}$ & $\text{mAP}_{m}$ & $\text{mAP}_{l}$ \\
    \cmidrule(rl){1-13}
    Baseline & 38.2 & 58.9 & 41.6 & 22.0 & 41.9 & 49.4 & 34.8 & 55.8 & 37.0 & 16.3 & 37.4 & 50.4 \\
    \textit{w.} Converse2D & 38.3 & 58.9 & 41.8 & 22.6 & 41.5 & 49.2 & 34.7 & 55.8 & 36.9 & 16.7 & 37.4 & 50.1 \\
    \cmidrule(rl){1-13}
    \textit{w.} WRC & 38.8 & \textbf{60.1} & 42.2 & \textbf{23.7} & 42.3 & 49.1 & 35.0 & \textbf{56.7} & 37.0 & \textbf{17.2} & 37.6 & 50.2 \\
    \bottomrule
\end{tabular}}
\end{center}
\end{table*}

\subsection{Applying WRC to Feature Pyramid Network}
Table~\ref{tab:Mask-RCNN} evaluates whether our feature upsampling operator benefits a standard detection and segmentation framework. Specially, we replace the default feature upsampling  operation in the Feature Pyramid Network (FPN) of Mask R-CNN~\cite{he2017mask} (typically interpolation-based sampling) with Converse2D and our WRC, respectively, while keeping the rest of the architecture and training protocol unchanged. 
As shown in Table~\ref{tab:Mask-RCNN}, both inverse-operator variants improve upon the baseline, and WRC yields the most consistent gains. In particular, WRC substantially increases $\text{mAP}_{50}$ for both detection and instance segmentation, indicating better localization and mask quality. The improvements are especially pronounced for \textbf{small objects} ($\text{mAP}_{s}$), suggesting that WRC better preserves fine spatial details during multi-scale feature upsampling, which is critical for detecting and segmenting small instances.

\begin{figure*}[t]
  \vspace{-2mm}
  \begin{center}
    \centerline{\includegraphics[width=\linewidth]{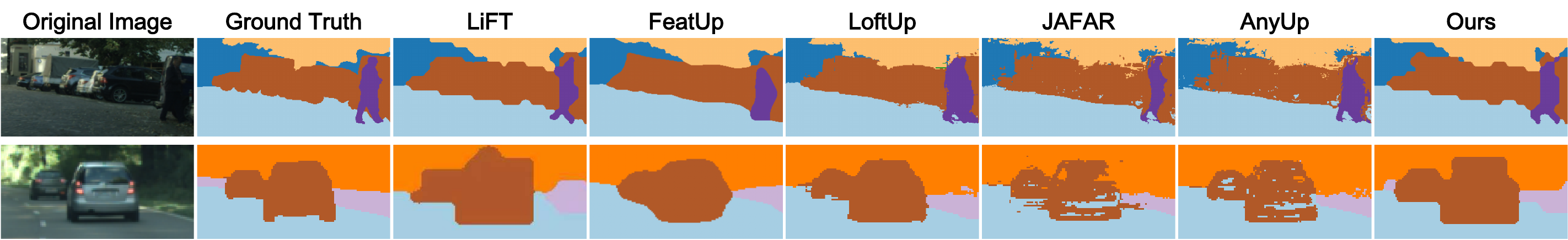}}
    \caption{Qualitative comparison of linear probing semantic segmentation on Cityscapes. We visualize predictions from representative methods (LiFT, FeatUp, LoftUp, JAFAR, AnyUp) against ground truth. Our method produces cleaner region boundaries and more coherent class assignments, and remains robust in challenging conditions such as low-light scenes.
    }
    \label{fig:visualization2}
    \vspace{-5.8mm}
  \end{center}
\end{figure*}

\subsection{More Qualitative Results}\label{sec}
\begin{figure*}[t]
  \vspace{-1mm}
  \begin{center}
    \centerline{\includegraphics[width=\linewidth]{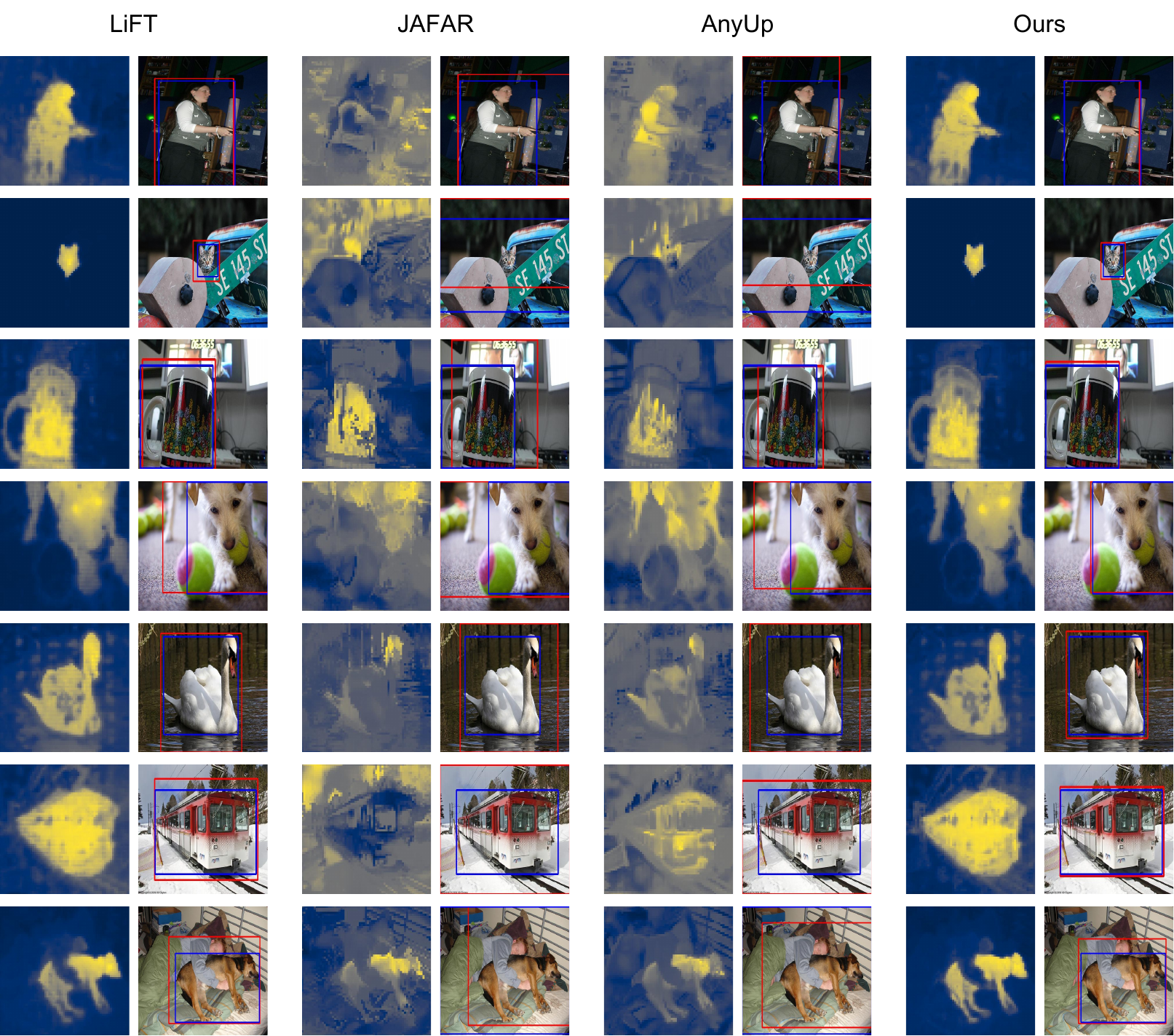}}
    \caption{Qualitative comparison of unsupervised object discovery on COCO20K. For each method, we show the eigen-attention (TokenCut) heatmap in the left panel and the resulting detection (predicted bounding box) in the right panel. \textcolor{blue}{Blue} and \textcolor{red}{red} boxes denote the ground-truth and predicted bounding boxes, respectively.}
    \label{fig:visualization3}
  \end{center}
\end{figure*}

Here we provide additional qualitative results on unsupervised object discovery on COCO20K. We apply TokenCut~\cite{tokencut}, which performs a graph cut on the feature affinity graph, to qualitatively assess different upsampling operators.
Figure~\ref{fig:visualization3} compares LiFT, JAFAR, AnyUp, and our method. For each method, we visualize the TokenCut heatmap (left) used to derive the object region, and the resulting predicted bounding box (right). Blue boxes denote ground truth, and red boxes denote predictions. Our method produces heatmaps that are more compact and object-centric, with clearer separation from the background. This leads to tighter and more accurate bounding boxes, particularly in challenging cases involving thin structures, cluttered backgrounds, or low-contrast objects. In contrast, competing upsamplers often yield diffuse responses, which can cause over-extended boxes or missed parts of the target.

\end{document}